# High Precision Medicine Bottles Vision Online Inspection System and Classification Based on Multi-Features and Ensemble Learning via Independence Test

MA Le, Wu Xiaoyue and LI Zhiwei*

*Abstract*: *To address the problem of online automatic inspection of drug liquid bottles in production line, an implantable visual inspection system is designed and the ensemble learning algorithm for detection is proposed based on multi-features fusion. A tunnel structure is designed for visual inspection system, which allows bottles inspection to be automated without changing original processes and devices. A high precision methods is proposed for vision detection of drug liquid bottles. Local background difference method is utilized as a soft on-off to capture bottle image. And an image gray level equalization preprocessing technology is used to eliminate the impact from illumination. Three features are designed, which contain blocked histogram of gradient,blocked histogram of gray and Raw-pixel. An ensemble learning algorithm is proposed based on independence test and multi-features fusion, after theoretically analysing the precondition of precision boosting to ensemble learning. Some result of analysis and comparison prove that the methods proposed is advanced compare with baseline methods.And specially, there exist a remarkable advantages in our method, when there are some noise in samples labels. We carried on a continuous test of 72 hours in practical production line, in which the error rate of inspection is less than 1‰ and performances of time and precision are superior to current manual detection. And hence the results of the test above prove that the visual inspection system designed and algorithm proposed are advanced and practical.*

## 1 Introduction

Drug safety is the critical issue of healthy and public livelihood. Accordingly, qualities of medicine bottles are the precondition to brug and the glass bottles effect liquid medicine specially. Compared with plastic glass bottle is easy to fragment, crack, deform and be polluted. In addition some impurity can drop into bottle in the productive process. Hence medicine bottles have to check strictly before leaving factories.

Although some technologies have been increasing such as industrial automation, pattern recognition, and artificial intelligence, inspection of medicine bottles the subdivision field of industry mostly still depends on manual works, which give rise to many problems including low efficiency, high fault and cost due to their subjectivity and tediousness of works. And meanwhile most of the quality problems are macroscopic, computer vision based technology is suited to online and automatic glass bottles inspect.

However, there are many challenges in visual inspection of glass medicine bottles, which are summarized as follows: 1) The types of defective medicine bottles are diverse and various, and defective appearances randomly on body of bottle; 2) it is imbalanced for data set due to low proportion of defective in total medicine bottles; 3）the glass material are transparent and reflective, and the phenomenon of light saturation thus may cover up defectives. The above problems restrict the practical application of computer vision technology in glass medicine bottle inspection.

The some current works related to the above problems are summarized as follows: In [4] image enhancement and morphological filtering methods are utilized to detect surface of anti-reflective glass materials under low-contrast. The surface defects of mobile phone glass screen are detected based on templates match and fuzzy C-mean cluster in [5]. In [6] the machine vision system is developed for fast detection of defects occurring on the surface of bottle caps by using sparse representation. PET bottle caps defect inspection is realized on based on localization matching, edge extraction and contrast detection in [7]. Hough transform is a common technology for shapes distinguishing, and in [8] an image processing application thus is implemented to detect and label faulty bottle packages for impeccable production by using Hough transform. In [9] Hough circles transform method is used to solve the problem of object location, and a framework is proposed by using of the saliency detection and template matching for bottle bottom inspection real time. This work is improved in [10], by using entropy rate superpixel circle detection and wavelet transform multi-scale filtering. And the texture area defect detection in bottle bottom was achieved. The authors report an automated system for inspecting foreign particles within ampoule injections in [11], in which multiple-features ranging from particle area, mean gray, geometric invariant moments, and wavelet packet energy spectrum are used in supervised learning to generate feature vectors. And they further improved this research and established an adaptive tracking method based on the adaptive local weighted-collaborative sparse model to achieve particle segmentation and tracking in [12] .

Although the existing works have achieved some results in different profiles, the following problems still exist: 1) existing design of inspection systems need to rebuild largely or even replace of the intrinsic production process with decreasing production rate and increasing costs; 2) Existing methods depend on relatively little data set, which are commonly less than 2000, and samples thus are not enough to truly cover all of possibility in production together with the problem of imbalance of data. Hence there exits a precision ceiling on these methods, which maybe not satisfy requirement of practical production; 3) The varieties of defective types in existing research work are relatively simple, and these methods may be ineffective to the problems containing multiple defect sub-classes; 4) In the most scenario the detected bottle is required to be fixed when it is shot to obtain image, so it's still a challenge to inspect bottles that move fast on the conveyor belt of production line.

To address these issues mentioned above, a monocular liquid medicine bottles visual online inspection system is designed in this paper, which is implantable without changing intrinsic production line. And a high precision detect algorithm based on ensemble learning with multi-features is proposed, which can be applied in a relatively large-scale dataset that can be 10,000 to 100,000 images. The main research contents in this paper are as follows:

1) An implantable visual inspection system is designed to detect moving medicine bottles and remove the defective products (see section 2 for details).

2) Some preprocessing technologies, including local background difference and gray mean value normalization, are utilized to capture image with bottle and to denoise medicine bottles images (See section 3 for details).

3) A multi-features extraction method are employed, which contains some improved features extraction technologies such as block histogram of gradient. And an ensemble learning method based on independence test is proposed to boost the precision of medicine bottle images detection for 10,000 to 100,000 images data with label noise(see section 3 and 4 for details).

And The main contribution of this paper is as follows:

1) The designed visual system only needs a one camera to inspect bottles, and the system has no any impact on the intrinsic production line. In addition the design of tunnel style structure and backlight projection reduces greatly the impact from external light.

2) The improved local background difference method is effective to capture the image with medicine bottle when bottles move on the production line, and the normalization method avoids the parameter adjustment problem which present to some traditional methods, and eliminates the influence of light from the plant and system light source.

3) The multi-features extraction methods can fully represent the essential information of the images, and have better ability to characterize various defective sub-classes. The proposed ensemble learning method break up the ceiling of detection precision. Moreover it's effective in the sub-classifiers screen and has strong tolerance capability for samples noise. Therefore, the method in this paper can realize high-precision online industrial detection of large-scale data.

## 2 Design of Visual Inspection System

The research of this article is drove by a actual project from a medical corporation, which need to make some automatic improvements for liquid medicine glass bottles inspection. As shown in Fig1, there are some defective bottles with various types of defects on the production line. And the most of defects for medicine bottles can be detected visually. In the past inspection depends completely on manual works, which thus cause the problems mentioned in the introduction. For this problem, we designed a vision-based automatic inspection system to replace manual.

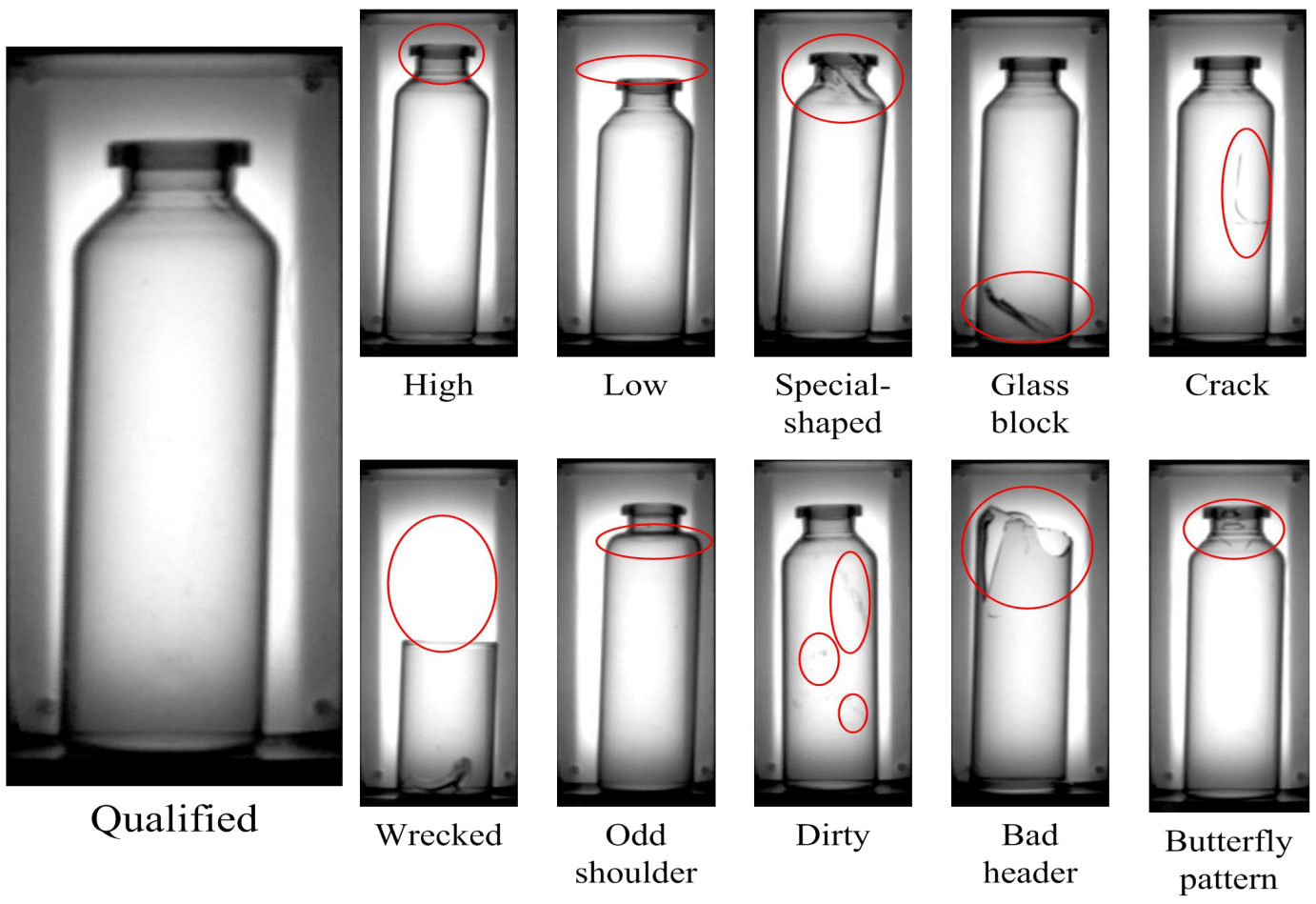

Fig1 The types of defective bottles

Fig 2 shows a schematic diagram of the designed monocular visual inspection system. The system is composed of industrial camera, light sources, darkroom, industrial control computers, switch modules, air pump, jet and power supply. Among these components, industrial cameras and light sources are mounted on both sides of the tunnel darkroom respectively.The darkroom is hollow, and the medicine bottles on transport disc can pass through the darkroom. When medicine bottle enters the dark room, the camera captures the image of the medicine bottle and transmits it to the industrial computer. And then the target detection programming in the industrial computer triggers the recognition command, and the system recognition programming makes some preprocesses and detects the current medicine bottle image. If the current image is detected to be qualified, the medicine bottle will pass through the dark room, otherwise the system will trigger the switch module to open the air pump jet, and the defective bottle is removed from production line.

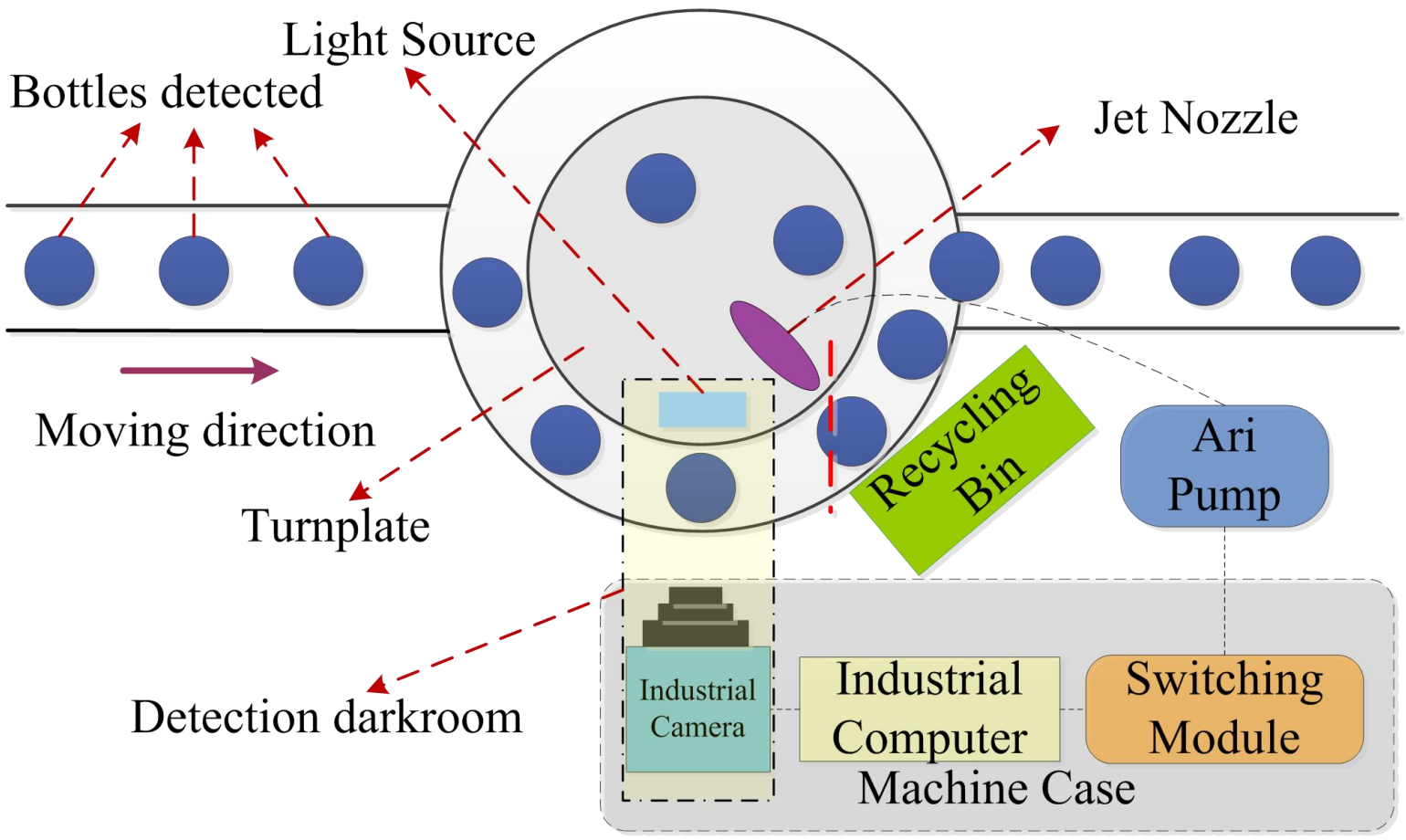

Fig2 The schematic diagram of monocular camera vision detect system designed

The defects of the medicine bottles can be observed horizontally. Therefore, different from vertical projection used in [13, 14], we employ the backlight horizontal projection to capture images of the medicine bottle. As shown in the side view of the darkroom of the detection system in Fig3, the homogenizing panel converts the light of the LED from the point light source to the surface light source, and the light projects it horizontally to the industrial camera. It's worth mentioning that although it may results in a certain amount of light saturation in the image in this way will, it effectively avoids the uncertain effects caused by reflections. Due to the size of bottle is constant, the parameters of camera including aperture, exposure time and light source brightness can be adjusted to achieve the most favorable imaging effect for detection and be fixed.

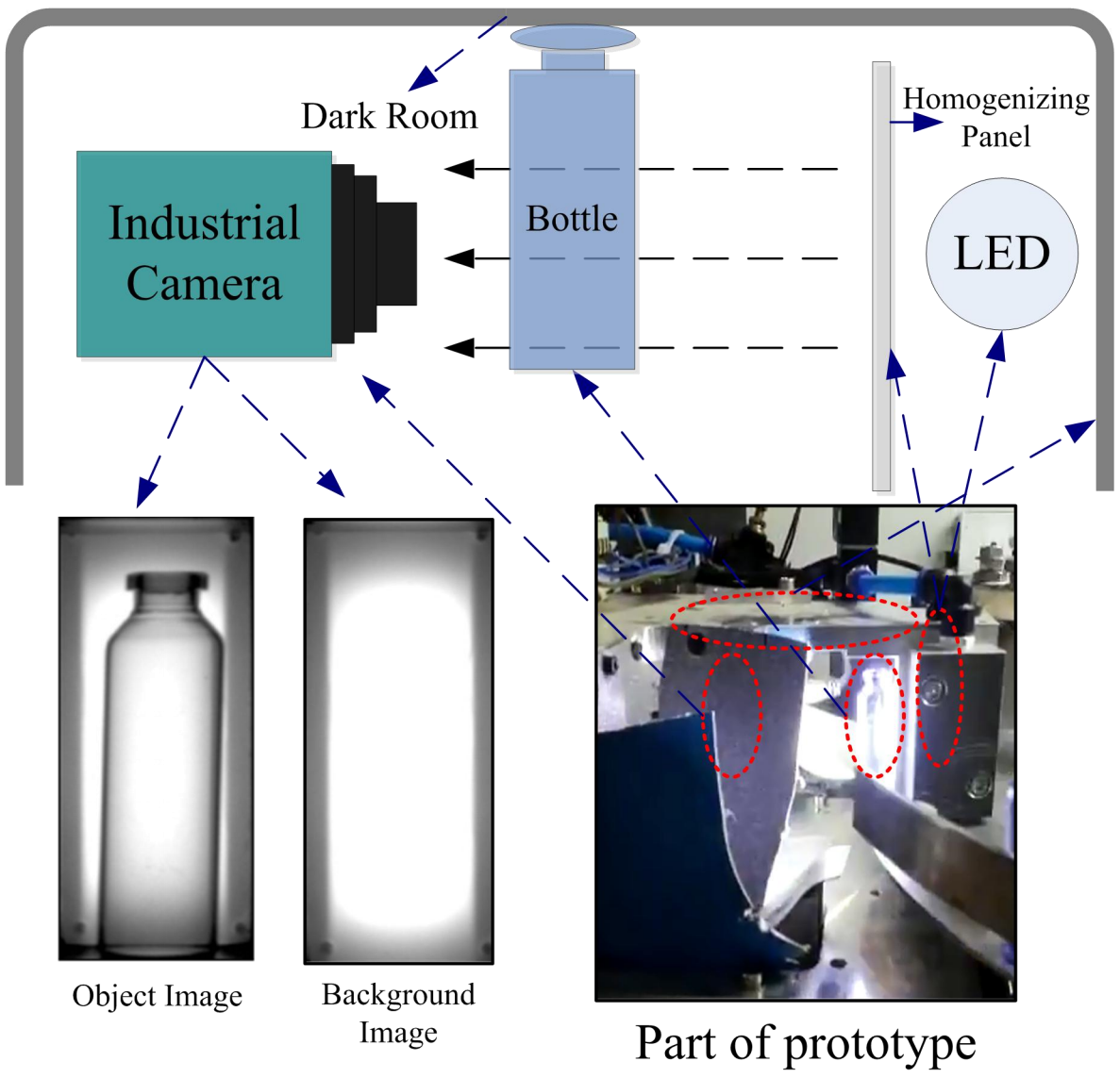

Fig3 The schematic diagram of horizontal backlit photography

## 3 Online Object Detection and Preprocessing

### 3.1 Local Background Difference

As described in the previous sections, the inspection system must first determine whether there exists the bottle on the image currently captured, and the image will be further processed if exists .Photoelectric switches are often employed in industrial production lines to provide detection signals, but this way is not very applicable for the working conditions described in this paper. There are twofold of reasons as follows: 1) due to the high speed movement of the medicine bottles (85-105 bottles/minute, frequency fluctuations will occur in processes such as refueling and stopping, and the average is about 95 bottles/minute), it is hard to set an appropriate time interval to trigger image capturing, which could cause the some uncertainties of the detection area position of medicine bottle in the image, and could thus make the processing difficult. 2) The reliability are relatively low for low price photoelectric switch, and meanwhile the coast is bound to increase when the nice module is adopted and assembled.

Hence, we make full use of the information of images, and then the technology called background difference is leveraged to capture the image with bottle. The basic principle is to estimate the energy difference between the background image and the current image detected. If the sum absolute energy difference exceeds some threshold value it will be identified that bottle presents at current image, and vice versa. In this paper, we improve the traditional background difference method according to actual situation.

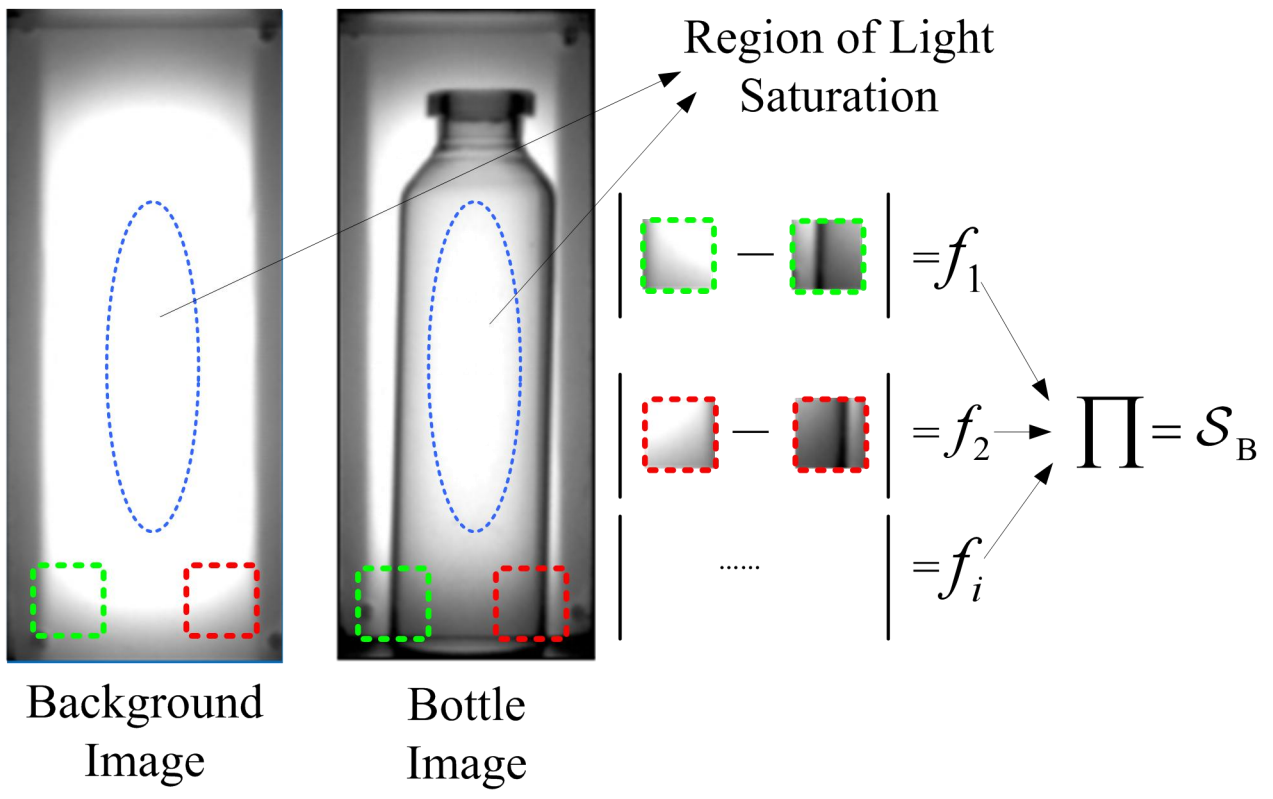

Fig4 The schematic of local background difference processing

As shown in Fig4, the pixel gray value of the background image $\mathbf{I}_B$ and the image with medicine bottle $\mathbf{I}_C$ are different, so the image with bottle can be identified by computing the sum absolute errors of pixel value between $\mathbf{I}_B$ and $\mathbf{I}_C$. It can be seen that the pixel difference of two images is very small in some regions, and hence it is unnecessary to computing the pixel difference of these regions. In this paper the we select several patches $\mathcal{P}_i$, and only compute the pixel difference in these patches. The detail operation can be described as:

$$\begin{cases} \mathcal{S}_B = \operatorname{sgn}\left(\prod_{i}^{N_{\mathcal{P}}} f\left(\mathbf{I}_B, \mathbf{I}_C \mid \mathcal{P}_i\right)\right) \\ f\left(\mathbf{I}_B, \mathbf{I}_C \mid \mathcal{P}\right) \\ = \mathcal{B}\left(\sum_{x,y\in\mathcal{P}} \left|\mathbf{I}_B(x,y) - \mathbf{I}_C(x,y)\right| - \theta_{\text{thres}}\right) \\ \mathcal{B}(x) = \begin{cases} 1, x > 0 \\ 0, x \le 0 \end{cases} \end{cases} \tag{1}$$

where $N_{\mathcal{P}}$ is number of the detection patch, $\theta_{\text{thres}}$ is the energy difference threshold. And $\mathcal{S}_B$ represents the identification result, that is if $\mathcal{S}_B = 1$ it means that there exits bottle, and vice versa.

Remark 1: The frame rate of the camera is 60pfs, and hence there may be continuous f rames of images that cam allow $\mathcal{S}_B = 1$. For this phenomenon, a flag variable is used to trig ger to processes programming only when $\mathcal{S}_B$ jumps from 0 to 1.

It can be seen from Fig4 and (1), we can obtain the image all the way, in which bottl e is in the centre, by configuring the position and size of $\mathcal{P}_i$. Considering the incompleteness at top of some defective images, the $\mathcal{P}_i$ is set at the bottom of the image.

The mean value image of multiple frames is used as the background image to eliminate noise, which can be expressed as:

$$\mathbf{I}_{\mathrm{B}}(x,y)=\frac{1}{N_{\mathrm{B}}}\sum_{i=1}^{N_{\mathrm{B}}}\mathbf{I}_i(x,y) \tag{2}$$

where $N_{\mathrm{B}}$ represents the number of image frames.

3.2 Normalization based on gray-scale mean value

Although light source is equipped the darkroom is not completely enclosed. And when bottle moves in and out darkroom, external illumination could impact on image quality. And hence image need to be normalized before subsequently working.

Gamma correction has been widely used for gray level transformation of image, however it is difficult to tune its parameter. At the same time, due to its nonlinear transformation, image distortion could be produced, which can conceal some defective information and generate pseudo defects. For this reason, the gray level equalization is designed to normalize the image (the current image pixel depth is 32-bit gray scale, ranging from 0 to 1). And the image normalized $\mathbf{I}_{\mathrm{N}}(x,y)$ can be obtained as follows:

$$\mathbf{I}_{\mathrm{N}}(x,y)=\frac{\lambda_{\mathrm{avg}}N_{\mathrm{avg}}}{\sum_{x,y\in D_{\mathbf{I}}}\mathbf{I}(x,y)}\mathbf{I}(x,y) \tag{3}$$

where $N_{\mathrm{avg}}$ represents the number of image pixels, $\lambda_{\mathrm{avg}}=0.5$ is the tunable factor. It can be seen from (3) that the total energy of the image normalized is a constant, which can eliminates the impact of illumination transformation effectively. In order to prevent the problem of parameters tuning, the $\lambda_{\mathrm{avg}}$ is generally set as a fixed constant value.

**4 Multi-features Extraction**

Due to variety and randomness of defective bottles, it's hard to characterize all inherent information of bottle images by using single feature. As a sequence, multi-features extraction technology is utilized to express fully the quality information of bottles, and the features contain block histogram of gradient, block gray value histogram and raw-pixels.

4.1 BHoG Feature Extraction

The feature of histogram of gradient (HoG) is effective and used widely in field of computer vision. however, defective positions on bottles present randomly, thus the traditional global HOG is not enough to describe defects of bottles well. Consequently we improve it into the block histogram of gradient (BHoG).

Sobel operator is used to compute gradients of image, and $\mathbf{G}_{\mathrm{x}}(x,y)$ and $\mathbf{G}_{\mathrm{y}}(x,y)$ denote gradient image of $\mathbf{I}(x,y)$ in x and y direction respectively, whose relation can be written:

$$\begin{cases}\mathbf{G}_{\mathrm{x}}(x,y)=\mathbf{T}_{\mathrm{x}}^{\mathrm{s}}*\mathbf{I}(\mathrm{x},\mathrm{y})\\ \mathbf{G}_{\mathrm{y}}(x,y)=\mathbf{T}_{\mathrm{y}}^{\mathrm{s}}*\mathbf{I}(\mathrm{x},\mathrm{y})\end{cases} \tag{4}$$

where $\mathbf{T}_{\mathrm{x}}^{\mathrm{s}}$ and $\mathbf{T}_{\mathrm{y}}^{\mathrm{s}}$ are the 3×3convolution template of Sobel operator in the x and y directions of respectively, and the parameters are arranged as follows:

$$\mathbf{T}_x^s=\begin{bmatrix}-1 & 0 & 1\\ -2 & 0 & 2\\ -1 & 0 & 1\end{bmatrix},\mathbf{T}_y^s=\begin{bmatrix}-1 & -2 & -1\\ 0 & 0 & 0\\ 1 & 2 & 1\end{bmatrix} \tag{5}$$

Polar coordinate transformation is employed to obtain gradient magnitude images $\mathcal{M}(x,y)$ and orientation image $\boldsymbol{\theta}(x,y)$:

$$\begin{cases}\mathcal{M}(x,y)=\sqrt{\mathbf{G}_x(x,y)+\mathbf{G}_y(x,y)}\\ \boldsymbol{\theta}(x,y)=\tan^{-1}\left(\dfrac{\mathbf{G}_y(x,y)}{\mathbf{G}_x(x,y)}\right)\end{cases} \tag{6}$$

And two of images $\mathcal{M}(x,y)$ and $\boldsymbol{\theta}(x,y)$ are divided into several regions denoted by $\langle \mathcal{C}_i^{\text{HoG}} \rangle$, and the orientation histogram feature $\mathbf{F}_i^{\text{HoG}}$ is formed from points within region $\langle \mathcal{C}_i^{\text{HoG}} \rangle$, as sh own in Fig.5.

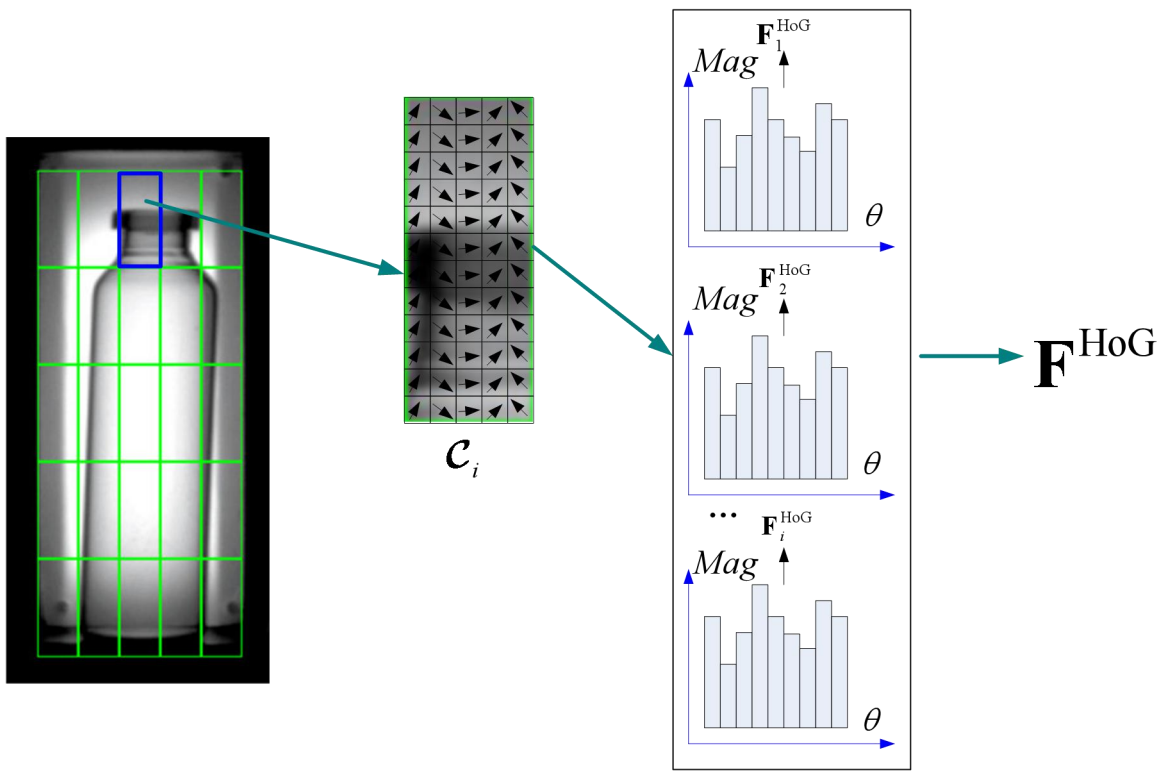

Fig5 The schematic of block histogram of gradient

Let the number of histogram bin be $N_{\text{bin}}^{\text{HoG}}$, then the HoG feature vector $\mathbf{F}_i^{\text{HoG}}$ in region $\mathcal{C}_i^{\text{HoG}}$ can be compute:

$$\begin{cases}\mathbf{F}_i^{\text{GoH}}=\sum\limits_{x,y\in\mathcal{C}_i^{\text{HoG}}}\boldsymbol{\gamma}_{\mathbf{s}}\left(\left\lfloor\dfrac{\boldsymbol{\theta}(x,y)}{\Delta_\theta}\right\rfloor\right)\mathcal{M}(x,y)\\ \boldsymbol{\gamma}_{\mathbf{s}}^j(x)=\begin{cases}0,\mathbf{s}_j\neq x\\ 1,\mathbf{s}_j=x\end{cases},\Delta_\theta=\dfrac{2\pi}{N_{\text{bin}}^{\text{GoH}}}\end{cases} \tag{7}$$

where $\mathbf{s}=[0,1,2,......,N_{\text{bin}}-1]^{\text{T}}$. And thus we combine all HoG feature vectors to form BHoG f eature vector of detection image.

4.2 BGH and RAW feature extraction

Gray value histogram is another simple and effective feature extraction technology. And i t is an especially effective expression for some random textures such as the stains on bottle. Same as BHoG, this feature is improved by blocking. The Image detected is divided into a set of regions that denote $\langle \mathcal{C}_i^{\text{GH}} \rangle$, and let the number of histogram bins be $N_{\text{bin}}^{\text{GH}}$,then the BG H feature $\mathbf{F}_i^{\text{GH}}$ in region $\mathcal{C}_i^{\text{GH}}$ can be obtained:

$$\mathbf{F}_i^{\text{GH}}=\sum_{x,y\in\mathcal{C}_i^{\text{GH}}}\boldsymbol{\gamma}_{\mathbf{s}}\left(\left\lfloor\frac{\mathbf{I}(x,y)}{\Delta\theta_{\text{GH}}}\right\rfloor\right),\Delta\theta_{\text{GH}}=\frac{256}{N_{\text{bin}}^{\text{GH}}} \tag{8}$$

where function vector $\boldsymbol{\gamma}_s$ is mentioned in equation (7).

Look-Up-Table(LUT) technology can be leveraged to accelerate computing equation(8)[16]. The gray values of image range from 0 to 255 and are all integer for 8bit image. The values of 8bit gray image constitute a finite set, and the bin index of every pixel value is fixed when the bin number $N_{\text{bin}}^{\text{GH}}$ of histogram is given. Therefore the indexes of all of pixel values can be calculated in advance according to $N_{\text{bin}}^{\text{GH}}$ . And $N_{\text{bin}}^{\text{GH}}$ can be mapped to table matrix, each column is a one-hot vector: $p, N_{\text{bin}}^{\text{GH}} \to \mathbf{V}$ . And hence we can fast obtain the index of histogram by look up the table matrix without calculating index repeatedly and equation (8) can be rewritten as:

$$\mathbf{F}_i^{\text{GH}} = \sum_{x,y \in \mathcal{C}_i} \mathbf{V}\left(\mathbf{I}(x,y), N_{\text{bin}}^{\text{GH}}\right) \quad (9).$$

For representing better some tiny defects such as subfissure and dust, we increase the RAW feature which is extracted by downsampling detected image directly. As shown in Fig 6, detected image is downsampled from size of $M \times N$ to $m \times n$, and the small image is reshaped into a $m \times n$ dimensional vector $\mathbf{F}^{\text{RAW}}$ that is just RAW feature. RAW greatly reduce dimension of the original image while maintaining all of tiny defects.

The above three features can characterize and cover all sample informations of different levels well, and subsequent detection and classification methods are exploited based on these features.

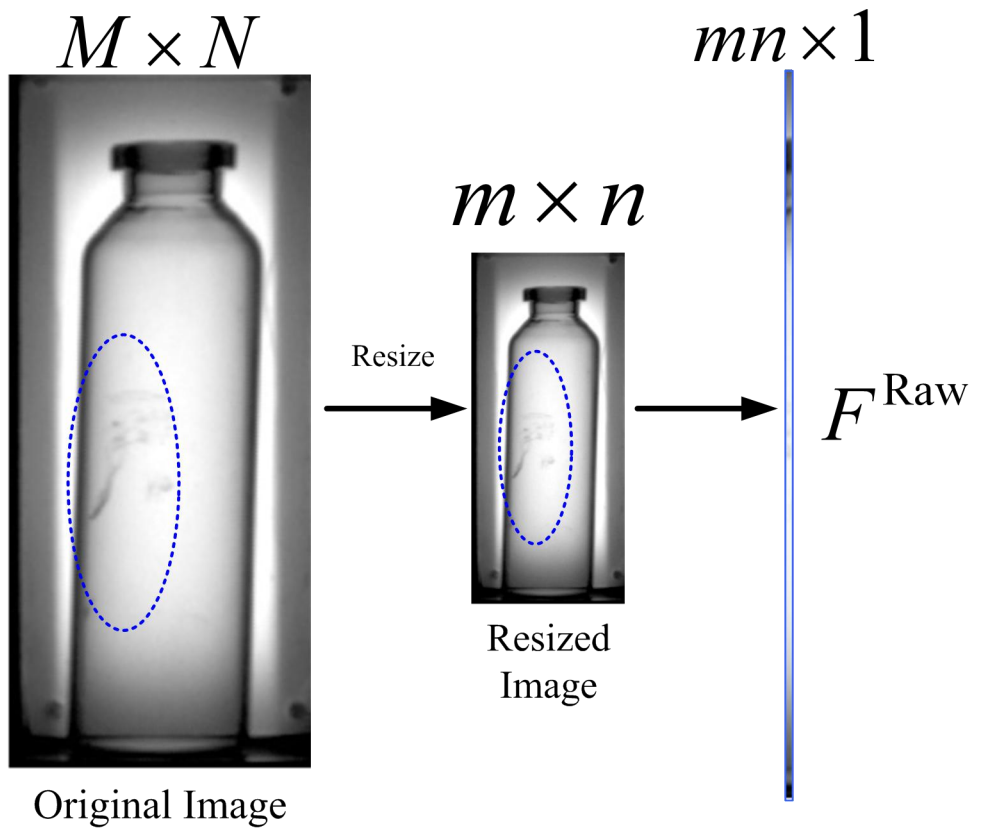

Fig6 The schematic of RAW feature

## 5 Ensemble learning for bottle inspection

### 5.1 Analysis for error rate of ensemble learning

We consider the problem of binary classification. If there exists a set of sub-classifiers, $h_i, i = 1,2,...,T$ , $T$ is odd number commonly, and for a sample $x$ the classification result can be expressed as $h_i(\mathbf{x}) \in \{-1,1\}$,where 1 denotes positive and -1 denotes negative, a combined type ensemble classifier can be obtained:

$$H(\mathbf{x}) = \text{sgn}\left(\sum_{i=1}^{T} h_i(\mathbf{x})\right) \quad (10)$$

Ensemble learning has the capability of accuracy boost by combining with a serial of sub-classifiers, which is especially suitable for some industrial scenes where high precision is required. However, it needs some preconditions for these sub-classifiers to boost precision, and hence we make some analysis.

Here, we assume that all of sub-classifiers are independent of each other and then have the same error rate that denotes $\varepsilon$ , that is:

$$P\big(h_i(\mathbf{x}) = y(\mathbf{x})\big) = 1-\varepsilon \tag{11}$$

where $y(\mathbf{x})$ is the the label of sample $\mathbf{x}$, then the precision of $H(x)$ mentioned in equation (10) is:

$$P\big(H(x) = y(x)\big) = 1-\sum_{k=0}^{\lfloor T/2 \rfloor}\binom{T}{k}(1-\varepsilon)^k \varepsilon^{T-k} \tag{12}$$

A cluster of curves can be plotted according to equation (12), which response the relations between precision of single classifier and the one of combined classifier under diverse $T$, as shown in figure 7. And figure 7 reveals that if precision of single classifier is greater than 50% the accuracy rate of combined classifier will increase with the increase both of precision of single classifier and the number of sub-classifier $T$.

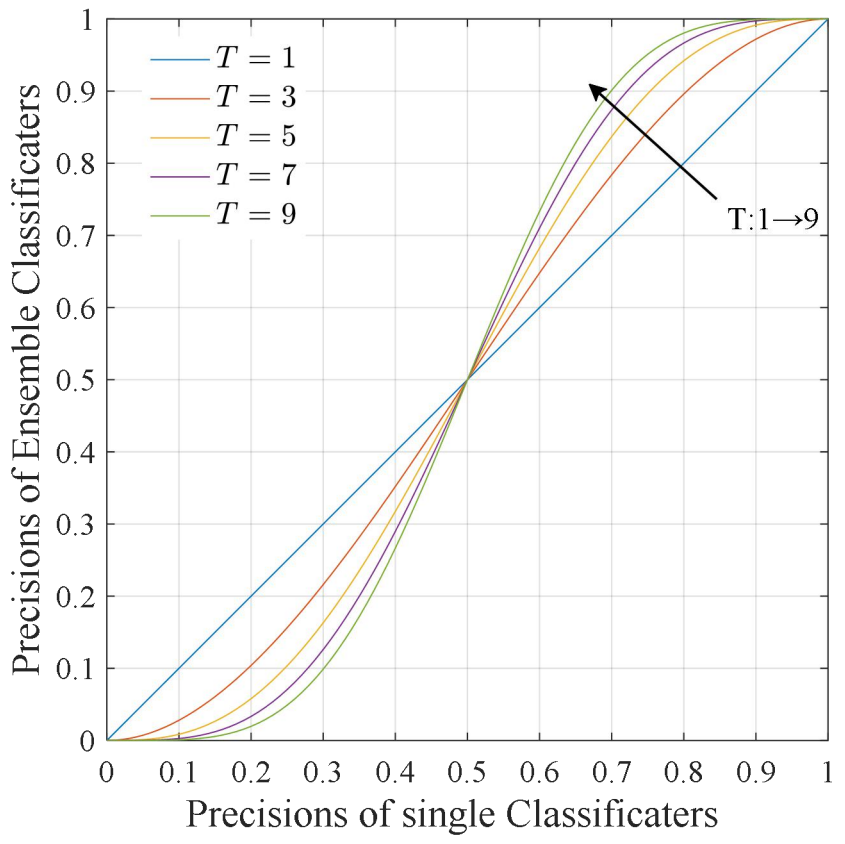

Fig7 The relation between number of classifier and precision

5.2 The ensemble method based on independence testing

The above analysis provides a theoretical support for improving the accuracy of classifier. However, the this conclusions have to have several preconditions. Firstly, the accuracy rate of a single classifier must be greater than 50%, which is relatively easy to be satisfied in practical applications. The more important precondition is the independence of sub-classifier. However, it's too hard to be meet in practice, especially when single classifier has a relative high precision.

Hence, we propose an ensemble learning algorithm based on the independence test, which is able to select some sub classifier that is independent of each other and holds relative high accuracy. Let the outputs of two sub classifiers, that denote A and B, be $h_{\mathrm{A}}$ and $h_{\mathrm{B}}$ respectively, if two of results are independent of each other then:

$$\begin{aligned} & P(h_{\mathrm{A}} \neq h_{\mathrm{B}}) \\ & = P(h_{\mathrm{A}} \neq y, h_{\mathrm{B}} = y) + P(h_{\mathrm{A}} = y, h_{\mathrm{B}} \neq y) \\ & = P(h_{\mathrm{A}} \neq y)P(h_{\mathrm{B}} = y) + P(h_{\mathrm{A}} = y)P(h_{\mathrm{B}} \neq y) \\ & \xrightarrow{P(h_* = y) \triangleq p_*^{y},\ P(h_* \neq y) \triangleq p_*^{\tilde{y}},\ * \in \{\mathrm{A},\mathrm{B}\}} \\ & P(h_{\mathrm{A}} \neq h_{\mathrm{B}}) = p_{\mathrm{A}}^{\tilde{y}} p_{\mathrm{B}}^{y} + p_{\mathrm{A}}^{y} p_{\mathrm{B}}^{\tilde{y}} \end{aligned} \tag{13}$$

And a random variable $X_{\mathrm{h}} \in \{0,1\}$ can be defined as follows:

$$\begin{cases} X_{\mathrm{h}}(h_{\mathrm{A}} \neq h_{\mathrm{B}}) = 1 \\ X_{\mathrm{h}}(h_{\mathrm{A}} = h_{\mathrm{B}}) = 0 \end{cases} \tag{14}$$

Then its expectation is:

$$
\begin{aligned}
&E(X_{\mathrm{h}}) = X_{\mathrm{h}}(h_{\mathrm{A}} \neq h_{\mathrm{B}})P(h_{\mathrm{A}} \neq h_{\mathrm{B}}) \\
&+ X_{\mathrm{h}}(h_{\mathrm{A}} = h_{\mathrm{B}})P(h_{\mathrm{A}} = h_{\mathrm{B}}) \\
&= p_{\mathrm{A}}^{\tilde{y}} p_{\mathrm{B}}^{y} + p_{\mathrm{A}}^{y} p_{\mathrm{B}}^{\tilde{y}}
\end{aligned} \tag{15}
$$

On the other hand, $E(X_{\mathrm{h}})$ can be estimated from a testing data set:

$$
\lim_{N\to\infty} \frac{1}{N} \sum_{i=1}^{N} X_{\mathrm{h}}^{i} = E(X_{\mathrm{h}}) \tag{16}
$$

where $N$ is the samples number of testing data set.

As a sequence, there exist the necessary condition of independence that equation (15) is equivalent to (16). And hence we propose the test algorithm to select a set of sub-classifier that are independent of each other on the basis of above condition.

Meanwhile, there are significant impacts on independence derived from the configuration of features and their classifiers, that is it's hard to ensure independence for two classification methods if they work on the same feature. For this reason, we combine diverse classifier with multi-features mentioned above, obtaining candidate sub-classifiers. Multi-features can represent various profiles of image information. In addition, any mechanism of classification method is different from each other. Thus, it's feasible to make a set of candidate classification methods by combining diverse classifiers with multi-features, which could meet the above independence requirements. All of process of the algorithm proposed is descried as table 1.

**Remark** 2: Theoretically if the estimation as mentioned in equation(16) is equal to expectation the samples number need to tend to $\infty$, and thus it is necessary to do some slack operation to solve this constraint, see table 1 for detail.

## 6 Analysis and test

In this section we make some tests to evaluate the performance of the proposed method, which contain 1) analysis of relation between performance and parameters and configurations of features as well as sub-classifiers, 2)the analysis of some impacts of parameters of ensemble method and noise of sample labels on precision, and 3) the comparison with existing methods to verify the advancement of our method. The configuration of relevant software and image data set is described as in table 2.

Table 2 Configuration table of test

| Notion | Description |
|---|---|
| Operating System | Windows7 |
| Image process library | OpenCV 2.4.12 |
| Image dataset | https://pan.baidu.com/s/125RkFjjzxrq scCuVgtXz2A Extraction code: ejn9 |

### 6.1 Analysis of relation between performance and parameters

In this paper, some classification methods including random forest(RF)[17], gradient boosted decision tree(GBDT)[18], support vector machine(SVM)[19] and k-nearest neighbor(kNN)[20] are selected as candidate sub-classifiers for ensemble learning we proposed.

Mentioned above, the parameters of features in the ensemble learning algorithm can be selected in the sub-classifiers pool. However, it will give rise to a low precision of classification, if some are not set properly. And hence, in this subsection we make some analysis of the relation between precision and key parameters and configuration of feature and classifier to refine the sub-classifiers pool. The size of positive and negative samples for training in tests of this subsection are 6829 and 5097 respectively (11926 in total ), and the size of positive and negative samples for testing are 6538 and 4743, respectively(11281 pictures in total).

Table 1 Ensemble Learning Algorithm based on Independence Test for sub-classifier

**Ensemble-Learning based on Independence Test—Main Algorithm**

**Inputs:** Dataset for training $\mathcal{D}$, number of sub-classifiers T,Threshold of independence test $\theta_{\mathrm{IT}}$, Features pool $\mathcal{F}$ ,Sub-classifiers set $\mathcal{H}$, Current number of sub-classifiers $n_{\mathrm{H}}=0$, Error upper and lower bound of sub-classifier $\delta_{\mathrm{low}},\delta_{\mathrm{up}}$, Maximum to select $n_{\max}^{\mathrm{pool}}=100$, Ratio for training and test extracting from Dataset $\alpha_{\mathrm{train}}$ and $\alpha_{\mathrm{test}}$, Ratio for independence test $\beta$, Set of sub-classification methods $\{<\mathbf{h}_i,\mathbf{F}_i>\}=\varnothing$,

**Output:** $\{<\mathbf{h}_i,\mathbf{F}_i>\}$

Counter $n \leftarrow 0$

while$\left(n < n_{\max}^{\mathrm{pool}}\right)$

  $n \leftarrow n+1$

  if $n_{\mathbf{H}} == T$ :

    break

  Randomly select $\mathbf{h}$ and $\mathbf{F}$ from $\mathcal{H}$ and $\mathcal{F}$

  $\mathbf{h},\mathbf{F},\delta_{\mathrm{false}},p_{\mathbf{h}}^{y},p_{\mathbf{h}}^{\tilde{y}} \leftarrow \mathrm{Train}\left(\mathbf{h},\mathbf{F},\mathcal{D},\alpha\right)$

  if not$(\delta_{\mathrm{low}} < \delta_{\mathrm{false}} < \delta_{\mathrm{up}})$

    continue

  if $n_{\mathrm{H}} == 0$ :

    $\{\langle\mathbf{h}_i,\mathbf{F}_i\rangle\} \leftarrow \mathbf{h},\mathbf{F}$

  else:

    Flag of independence test $s_{\mathrm{flag}} \leftarrow 1$

    for $i:1 \rightarrow n_{\mathrm{H}}$

      $\mathcal{D}_{\mathrm{IT}} \leftarrow$ Extract Samples set from $\mathcal{D}$ by ratio of $\beta$

      $S_{\mathrm{flag}} \leftarrow IT\left(\mathbf{h}_i,\mathbf{F}_i,\mathbf{h},\mathbf{F},\mathcal{D}_{\mathrm{IT}}\right)$

      if $S_{\mathrm{flag}} == 0$

        break

    if $S_{\mathrm{flag}} == 1$

      $\{\langle\mathbf{h}_i,\mathbf{F}_i\rangle\} \leftarrow \mathbf{h},\mathbf{F}$

      $n_{\mathrm{H}} \leftarrow n_{\mathrm{H}}+1$

return $\{\langle\mathbf{h}_i,\mathbf{F}_i\rangle\}$

**Train— Training algorithm of sub-classifier**

$\mathcal{D}_{\mathrm{train}},\mathcal{D}_{\mathrm{test}} \leftarrow$ Extract Samples set from $\mathcal{D}$ by ratio of $\alpha_{\mathrm{train}},\alpha_{\mathrm{test}}$

$\mathbf{h} \leftarrow$ Train Classifier on $\mathcal{D}_{\mathrm{train}}$

$\delta_{\mathrm{false}},p_{\mathbf{h}}^{y}p_{\mathbf{h}}^{\tilde{y}} \leftarrow$ Compute according to(13)

return $\mathbf{h},\mathbf{F},\delta_{\mathrm{false}},p_{\mathbf{h}}^{y},p_{\mathbf{h}}^{\tilde{y}}$

**IT— Algorithm of independence test**

$E(\mathbf{X}_h) \leftarrow$ Compute according to (16), $\mathbf{h}_i,\mathbf{F}_i,\mathbf{h},\mathbf{F}$ and $\mathcal{D}_{\mathrm{IT}}$

if $\left|p_{\mathbf{h}}^{y}p_{\mathbf{h}_i}^{\tilde{y}}+p_{\mathbf{h}}^{\tilde{y}}p_{\mathbf{h}_i}^{y}-E(X_h)\right|<\theta_{\mathrm{IT}}$ return 1

return 0

Firstly, the performances with respect to BHoG features with diverse parameters (rows and column of blocks ) and classifiers integrated are tested. we choose the same size of rows and column in each of test, that is the parameters setting are $5\times5$, $6\times6$, $7\times7$ etc. As shown in figure8, although there are different precision for different classifiers, the peak values of precision for all classifiers are almost distributed at the parameter $11\times11$. As a consequence, we set the parameters of rows and column of BHoG features blocks, which are in the classification pool ,to be around $11\times11$.

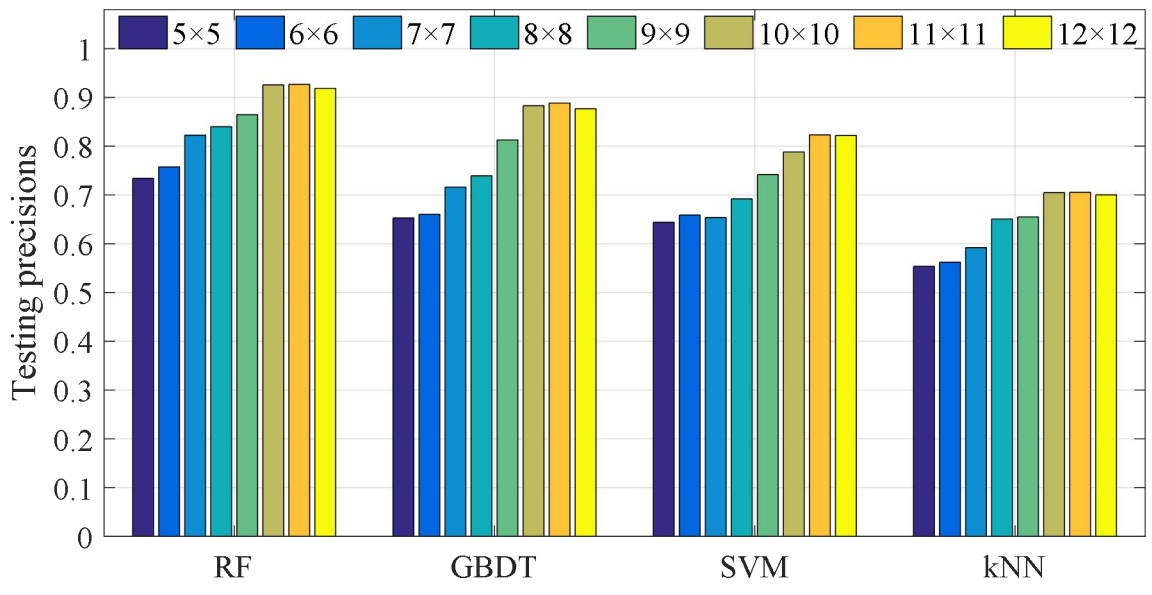

Fig8 Relation between precision and number of row and column of BHoG feature

Then the performances with respect to BGH features with diverse parameters (rows and column of blocks ) and classifiers integrated are tested, and meanwhile we get a similar result as shown in figure 9. There are different precision for different classifiers, the precision of all classifiers peak around $10\times10$. As a consequence, we set the parameters of rows and column of BGH features blocks to be around $10\times10$.

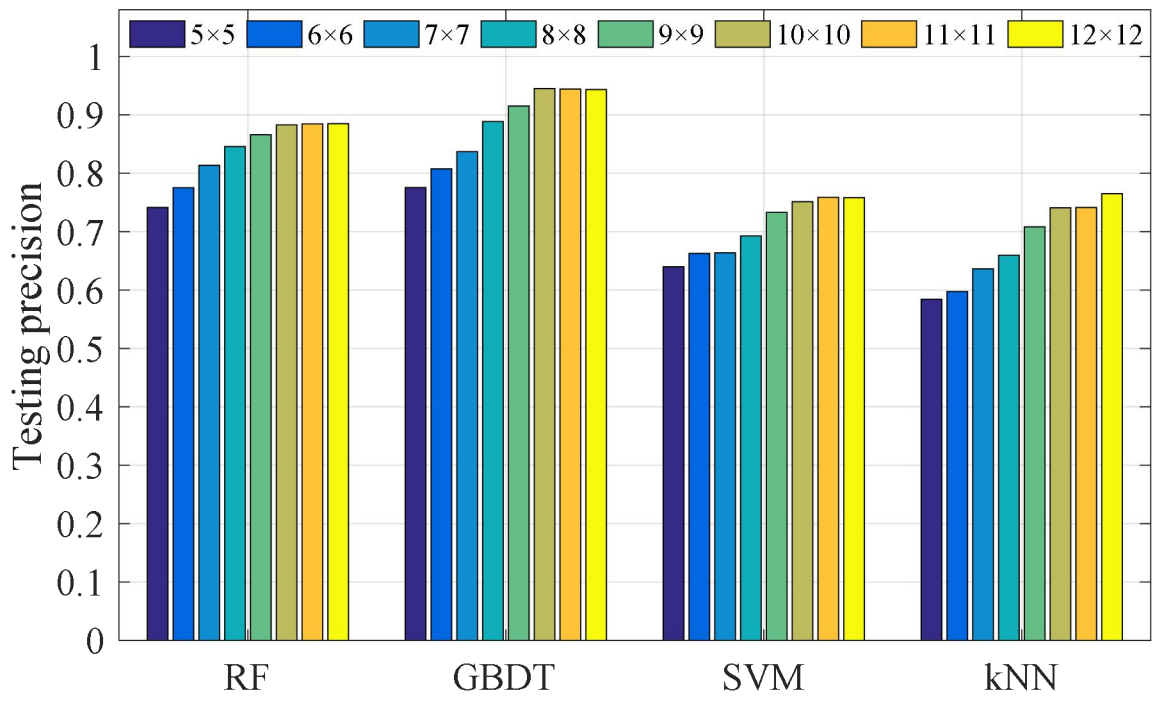

Fig9 Relation between precision and number of row and column of BGH feature

Finally, the influences on classification precision in different scale factors of RAW feature and classifiers are analyzed. The width and height of region of interest (ROI) of image are 150 and 356 respectively, and the tested scale factors are set as 0.2, 0.4, 0.6, 0.8, 1.0 respectively. As shown in figure 10, for all of tested parameters the precision SVM based classifier are highest overall, and there are similar precision in RF and GBDT classification. And the lowest precision come from kNN. Meanwhile, the peak values of all classifier are distributed around 0.6, and we select sub-classifiers for candidate pool according to this distribution law.

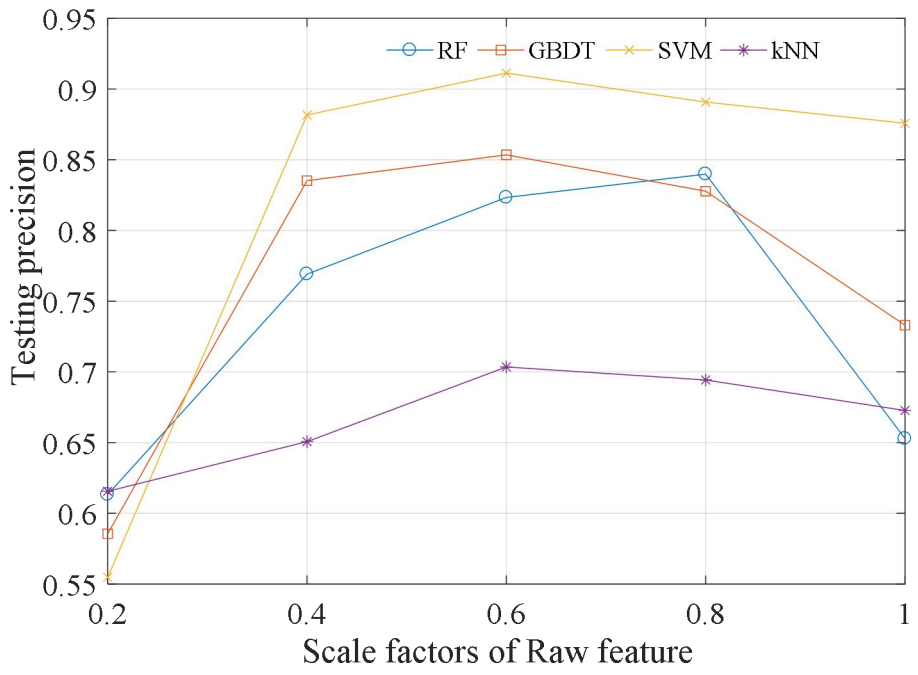

Fig10 Relation between precision and ratio of RAW feature

6.2 Classification methods analysis

In this subsection we make some analysis of performance of the ensemble classification method proposed. Firstly, we analyze the relation between testing precision and training time for different number of sub-classifiers. The size of samples for training is same as above subsection, and the number of positive and negative samples for testing of this subsection are 6569 and 6270 respectively (12839in total ).

Their relations are shown in table 3, which reveals that with increase of sub-classifiers number the precision don't improves obviously when number is bigger than 7, but the training time increases sharply. Hence we set number of sub-classifiers of the algorithm to be 7 in follow-up practical application.

Table 3 Relationship between number of sub-classifiers and precision and time performance

| Number of sub-classifiers | T=3 | T=5 | T=7 | T=9 | T=11 |
|---|---|---|---|---|---|
| Error rate(‰) | 30.77 | 12.15 | 0.55 | 0.47 | 0.47 |
| Ratio of training time | 1.00 | 3.56 | 12.03 | 50.94 | 190.32 |

There have common been some sample labels noise due to manual, and it's necessary to analyze the classification performance of the method proposed with sample labels noise. Hence we mix different ratios of incorrect labeled samples into training samples and test the robustness of our methods to noise of samples.

The relations between precision and ratios of sample labels are shown in figure 11, which indicate that sample labels noise has a little impact on precision when its ratio doesn't exceed 16% but the precision degrades sharply with the increase of ratio of the labels noise. The reasons we analyze lie in that 1) for sub-classifier training only a part of training samples data set were used, and thus the absolute number of noise labels was relatively low, which doesn't impact on the precision of sub-classifiers. 2) Some sub-classification methods and the ensemble learning algorithm proposed have the fault-tolerant capability for sample noise. 3) When the ratio of incorrect sample labels is relatively high, the guarantee of relatively high precision for each sub-classifier will be destroyed and thus it's hard to satisfy high precision of the ensemble classification.

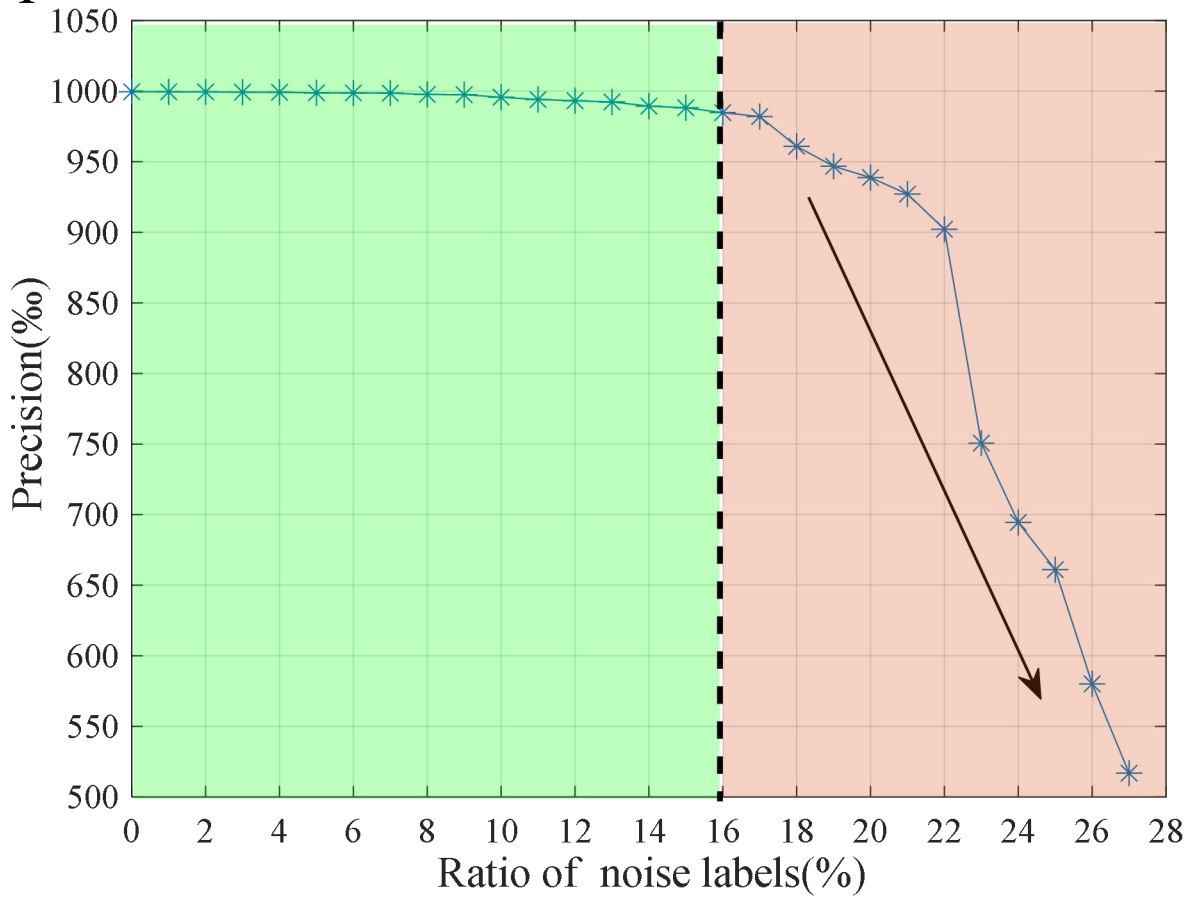

Fig11 Relations between noise ratio of labels and precision

Finally, we compare our algorithm with some existing methods to evaluate its performance. In this test the number of positive and negative samples, in which 3% of incorrect labels are mix up, for training are 12851 and 10331(total 23182), and the number of positive and negative samples for testing are 8085 and 7609 (total 15694).

Remark3: Although the defective products are much less than the qualified in practical industry, in consideration of algorithm generality and comparison the proportion of positive and negative samples for testing is set to 1:1 approximately.

The results of comparison are shown in table 4, which indicates that the performances of ensemble method proposed in this paper are superior to other methods compared in false positive rate and false negatives rate. There are some reasons for this results, which are as follows: 1) the features employed in [21] has a limited capability to represent all of defective types of bottles. 2) There are many types and random positions of defect on bottles, and thus it's hard to process these with high accuracy by using limited number of matching templates as described in [9]. 3) Although the method based on convolutional neural network(CNN) in [22] has relatively high accuracy, the labels noise may impact on performance. 4) Finally as mentioned above there two of advantages for our method that are the capability of representation to all defects depending on multi-features and fault-tolerant depending on ensemble based method, which can be a complementary reason for this comparison results. Therefore, our methods have some advancements to the problem of medicine bottle defects inspection.

Table 4 Detect performances comparisons of some methods to drug bottles data set

| **Methods** | **False positive number** | **False negatives number** | **False positive rate(‰)** | **False negatives rate(‰)** | **Precision (‰)** |
|---|---|---|---|---|---|
| HoG+SVM[21] | 375 | 362 | 46.38 | 47.58 | 953.04 |
| Templates Matching[9] | 1283 | 683 | 158.69 | 89.76 | 874.73 |
| CNN[22] | 89 | 261 | 11.01 | 34.30 | 977.70 |
| **Ours** | **6** | **4** | **0.74** | **0.53** | **999.36** |

## 7 Experiments

In previous section, we analyzed the performance of the methods in this paper, and verified the effectiveness and advancement of the methods for the problem of medicine bottles inspection. In this section, we describe the experiments to verify the performance of the methods and the system in the practical production line. The prototype machine setup of experiment is shown in figure 12 and some configurations for test are described in table 5.

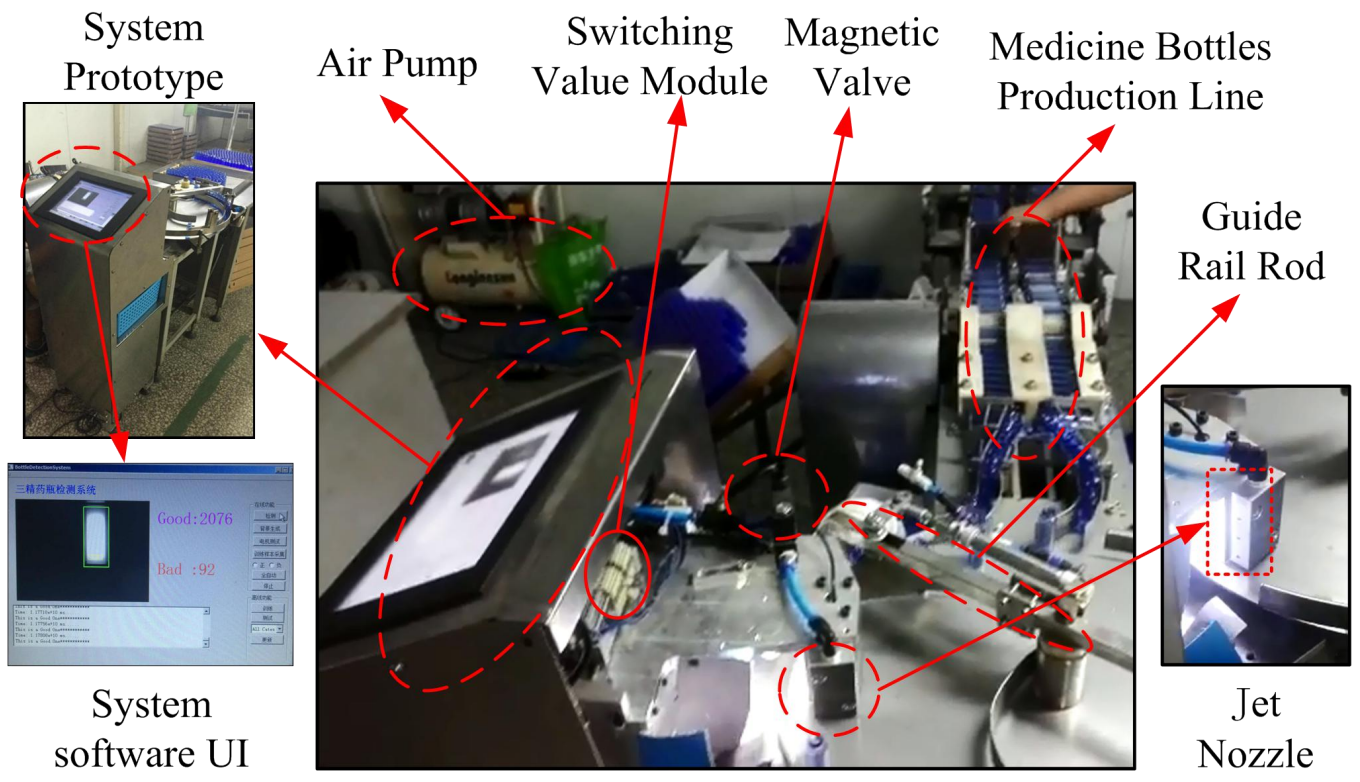

Fig12 Prototype machine setup of experiment

Table 5 Configuration of prototype machine

| **Notion** | **Description** |
|---|---|
| Industrial PC | Advantech AIMB-275L i7-7700 4Core 3.6GHz 16GB |
| Camera | JingHang JHSM130Bs USB Black and white, Distortionless and fixed-focus len |
| Light source | White 200mm×200mm 12V 10W |
| Switching Value Module | 8Channels 232-Serial 10A 7-24V |
| Software framework | Qt5.9.3 with VS2013 |

| ROI Parameters | x: 405, y: 39, width: 150, height: 356 |
|---|---|
| Difference region | [430,320,26,30], [460,320,26,30], [490,320,26,30] |

We have made 72 hours of continuous performance test to verify the effectiveness and practicability of the method and system. (It takes 8 hours shift work all day long in the factory of experiment, and thus it's enough for evaluation by running of system in 72 hours.)

The experimental results of test are shown in table 6, form which are drawn that are as follow: 1) The accuracy of object detection is the precondition and guarantee, and the error rate is very low due to the suitable configuration of regions, which provides the strong support for subsequent defect detection. 2) For the data of hundred thousands level and long time continuous running, total error rate is 0.727‰, which indicates that the algorithm and system in this paper have high accuracy and stability. 3) Although there was the imbalance between defected and qualified samples in practical production, system still held a considerable low false negatives rate which even was lower than false positive rate. Due to recycle for glass material, it's preferred to see that there exist a relatively low false negatives rate, which better fulfills production requirements.

Table 6 Results of practical experiment in 72 hours

| **Notion** | **Values** | **Notion** | **Value(‰)** |
|---|---|---|---|
| total number of samples | 413763 | Total error rate | 0.727 |
| False number of objective detection | 2 | Error rate of objective detection | 0.005 |
| False positive number | 298 | False positive rate | 0.730 |
| False negative number | 3 | False negative rate | 0.567 |

In the last of this section we compared the system with manual with respect to the performance of speed and accuracy of detection. For manual detection, there are 46081 bottles that had been detected by a worker in 8 hours. The worker is a team leader, and thus her professional qualification is not lower than the mean level of all workers.

The results of comparisons are show in figure 13. From sub-figure (a) is can be seen that the detection time of system is 82.69% blew the manual, and hence the speed of system is markedly higher than manual. And from sub-figure (b) is can be seen that although the worker holds a relatively high detection accuracy, the precision of our system is still higher than worker. The reasons lie in as follows: 1) it's hard for manual to detect some types of defects such as glass block and hidden crack. 2) Long time working results in exhaustion and decrease of attention of worker, which affects the detection precision of worker, but never impacts on machine system. 3) The algorithm and the prototype of system we set up has some advanced speciality to stability and accuracy. Therefore it is concluded that our system have some advancement in stability, speedability and accuracy compared with manual.

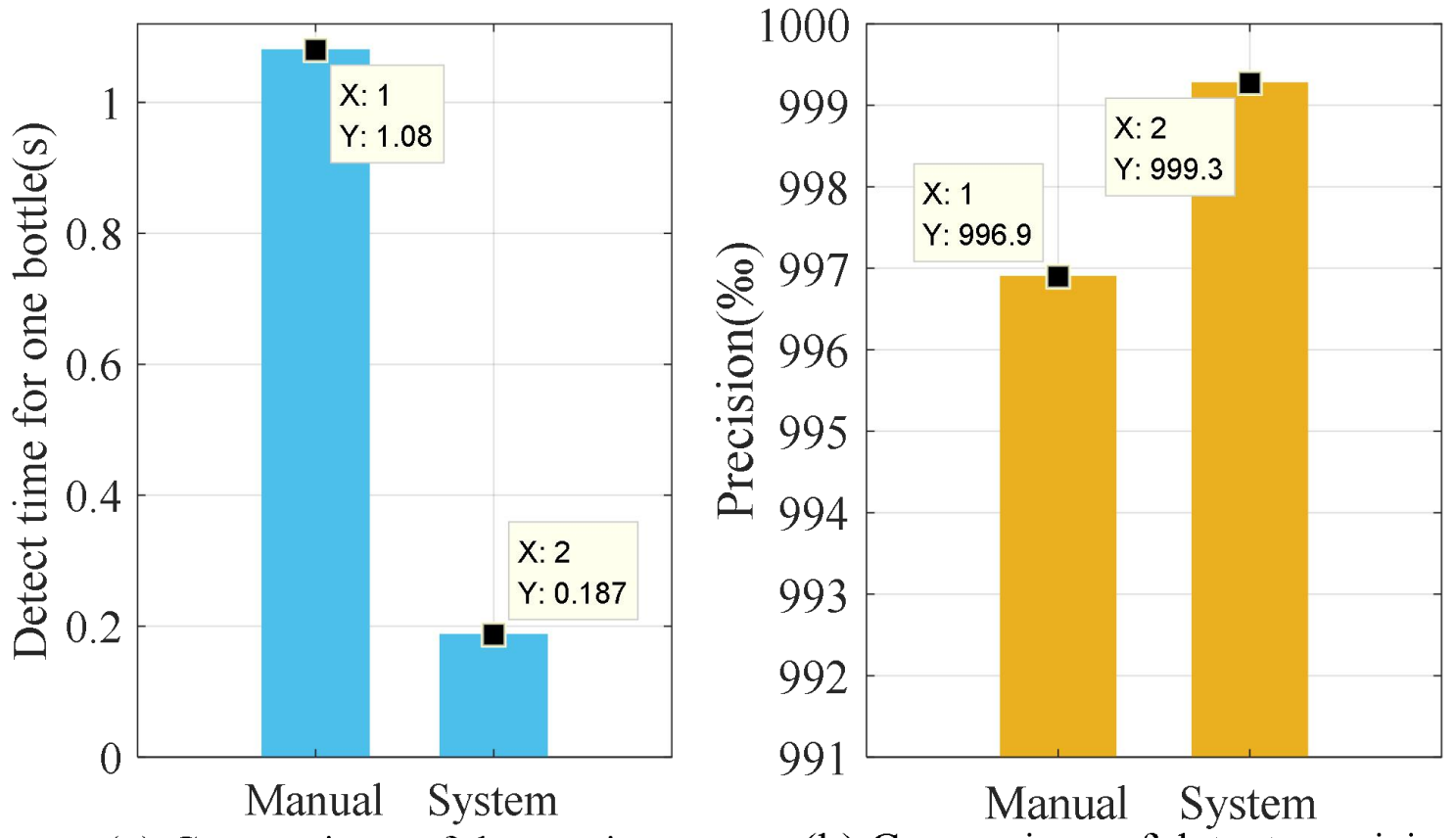

(a) Comparison of detect time (b) Comparison of detect precision

Fig13 The performance comparison between detect system and manual

## 7 Conclusions and future works

In this paper, we have presented a high precision “tunnel” style structural monocular vis ion online medicine bottle defects inspection system, and the related detection methods based on multi-features and ensemble learning classification via independent testing. There are som e conclusions which can be indicated as follows: 1) The prototype presented can run in prac tical production line without changing any original process and equipment. 2) The classificati on method proposed has high accuracy for the problem of medicine bottle defects detection due to multi-features extraction and ensemble learning methods by using independence testing technology. Specially anther strength of this method is that it has advantage when there so me noise in training sample labels. 3) Our system has long time and stably run with high p recision for the hundred thousands level data and has some advancements in speed and accu racy compared with the skilled worker.

In future we will investigate in some aspect as follows: 1) To expand to multi-classifica tion for some defective types. 2) To explore some methods to autonomously find new defect ive types that never have been seen. 3) To rewrite our detection program in some open sour ce operating system.